\documentclass{article}

\usepackage[nonatbib,final]{neurips_2021}

\usepackage{graphicx}
\graphicspath{{../../figs/}}

\usepackage{pdfpages}

\usepackage[utf8]{inputenc} 
\usepackage[T1]{fontenc}    
\usepackage{hyperref}       
\usepackage{url}            
\usepackage{booktabs}       
\usepackage{amsfonts}       
\usepackage{nicefrac}       
\usepackage{microtype}      

\usepackage{amsmath}
\usepackage{amssymb}
\usepackage{mathtools}

\DeclareMathOperator{\diag}{diag}
\def\given{\mid}

\usepackage{multirow}

\def\eg{e.g.,~}
\def\ie{i.e.,~}

\usepackage{bm}
\usepackage{lmodern}

\def\vec#1{{\bm #1}}
\def\diag{\text{diag}}
\def\mat#1{\mathbf{#1}}

\def\ErdosRenyi{{Erd\H{o}s-R\'{e}nyi}\xspace}

\usepackage{lmodern}
\usepackage{xspace}
\def\method#1{\texttt{#1}\xspace}%
\def\w2v{\texttt{word2vec}\xspace}%
\def\r2v{\texttt{residual2vec}\xspace}%
\def\n2v{\texttt{node2vec}\xspace}%
\def\DeepWalk{\texttt{DeepWalk}\xspace}%
\def\GAT{\texttt{GAT}\xspace}%
\def\FairWalk{\texttt{FairWalk}\xspace}%
\def\Glove{\texttt{Glove}\xspace}%
\def\LEM{\texttt{LEM}\xspace}%
\def\GCN{\texttt{GCN}\xspace}%
\def\GraphSAGE{\texttt{GraphSAGE}\xspace}%
\def\ErdosRenyi{{Erd\H{o}s-R\'{e}nyi}\xspace}

\usepackage{jabbrv}
\usepackage{cite}

\usepackage{url}

\makeatletter
\renewcommand\subsubsection{\@startsection{subsubsection}{3}{\z@}%
                                     {-0.5ex\@plus -1ex \@minus -.2ex}%
                                     {-1.5ex \@plus -.2ex}
                                     {\normalfont\normalsize\bfseries}}
\makeatother

\title{Residual2Vec: Debiasing graph embedding with random graphs}

\author{%
  Sadamori Kojaku$ ^1$, Jisung Yoon$ ^2$, Isabel Constantino$ ^3$, Yong-Yeol Ahn$ ^{1,3,4}$
  \And
  \\ \vspace{-1em}
  \begin{minipage}{17em}
    \centering
     $ ^1$Center for Complex Networks and Systems Research,
     Luddy School of Informatics, Computing and Engineering,
     Indiana University, USA
  \end{minipage}
  \And
  \\\vspace{-1em}
  \begin{minipage}{15em}
    \centering
    $ ^2$Department of Industrial Management and Engineering
    Pohang University of Science and Technology, South Korea
  \end{minipage}
  \And
  \\ \vspace{-1em}
  \begin{minipage}{12em}
    \centering
    $ ^3$Indiana University Network Science Institute
    Indiana University, USA
  \end{minipage}
  \And
  \\ \vspace{-1em}
  \begin{minipage}{12em}
    \centering
    $ ^4$Connection Science
    Massachusetts Institute of Technology, USA
  \end{minipage}
  \And
  \\ \vspace{-1em}
  \begin{minipage}{\hsize}
    \centering
    \texttt{\{skojaku, imconsta, yyahn\}@iu.edu, jisung.yoon92@gmail.com}
  \end{minipage}
}

\begin{document}

\maketitle

\begin{abstract}
    Graph embedding maps a graph into a convenient vector-space representation for graph analysis and machine learning applications.
    Many graph embedding methods hinge on a sampling of context nodes based on random walks.
    However, random walks can be a biased sampler due to the structural properties of graphs.
    Most notably, random walks are biased by the degree of each node, where a node is sampled proportionally to its degree.
    The implication of such biases has not been clear, particularly in the context of graph representation learning.
    Here, we investigate the impact of the random walks' bias on graph embedding and propose {\it residual2vec}, a general graph embedding method that can debias various structural biases in graphs by using random graphs.
    We demonstrate that this debiasing not only improves link prediction and clustering performance but also allows us to explicitly model salient structural properties in graph embedding.
\end{abstract}

\section{Introduction}
\label{sec:intro}

On average, your friends tend to be more popular than you.
This is a mathematical necessity known as the \textit{friendship paradox}, which arises due to a sampling bias, \ie
popular people have many friends and thus are likely to be on your friend list~\cite{Feld1991}.
Beyond being a fun trivia, the friendship paradox is a fundamental property of graphs: following an edge is a biased sampling that preferentially samples nodes based on nodes' degree (\ie the number of neighbors).
The fact that random walk is used as the default sampling paradigm across many graph embedding methods raises important questions: what are the implications of this sampling bias in graph embedding? If it is undesirable, how can we debias it?

\begin{figure}
    \centering
    \includegraphics[width=0.75\hsize]{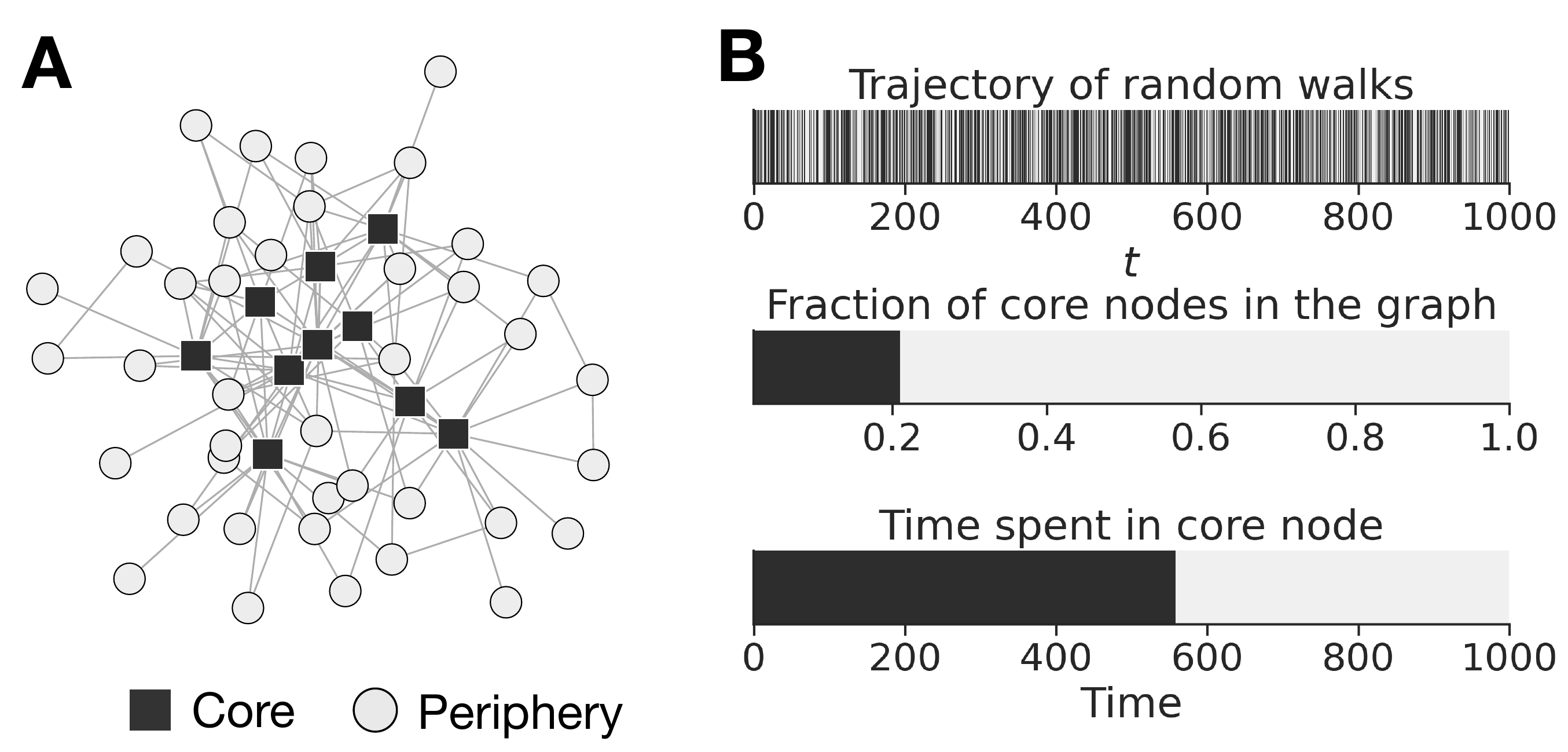}
    \caption{
        Random walks have a strong preference towards hubs.
        ({\bf A}) A toy graph generated by a stochastic block model with a core-periphery structure, where core nodes have more neighbors than peripheral nodes~\cite{Zhang2015}.
        ({\bf B}) Random walkers preferentially visit nodes with many neighbors, generating a trajectory that overrepresents the core nodes.
    }
    \label{fig:bias-in-random-walk}
\end{figure}

Graph embedding maps a graph into a dense vector representation, enabling a direct application of many machine learning algorithms to graph analysis~\cite{Chai2018}.
A widely used framework is to turn a graph into a ``sentence of nodes'' and then feed the sentence to \w2v~\cite{Perozzi2014, Grover2016,Dong2017, Mikolov2013}.
A crucial difference from word embedding is that, rather than using given sentences, graph embedding methods generate synthetic ``sentences'' from a given graph.
In other words, the generation of synthetic ``sentences'' in graph is an implicit modeling decision~\cite{Eriksson2021-kf}, which most graph embedding methods take for granted.
A common approach for generating sentences from a graph is based on random walks, which randomly traverse nodes by following edges.
The friendship paradox comes into play when a walker follows an edge (e.g., friendship tie): it is more likely to visit a node with many neighbors (e.g., popular individual).
As an example, consider a graph with a core-periphery structure, where core nodes have more neighbors than periphery (Fig.~\ref{fig:bias-in-random-walk}A).
Although core nodes are the minority, they become the majority in the sentences generated by random walks (Fig.~\ref{fig:bias-in-random-walk}B).
This is because core nodes have more neighbors than periphery and thus are likely to be a neighbor of other nodes, which is a manifestation of the friendship paradox.
Then, how does the sampling bias affect the embedding?

Previous approaches to mitigate the degree bias in embedding are based on modifying random walks or the post-transformation of the embedding \cite{Andersen2006-ki,Wu2019-qq,Tang2020-ta,Ahmed2020-vl,Ha2020-fs,Fountoulakis2019-fv,Palowitch2020-tu,Faerman2018-pd,Rahman2019-uh}.
Here we show that word2vec by itself has an implicit bias arising from the optimization algorithm---skip-gram negative sampling (SGNS)---which happens to negate the bias due to the friendship paradox.
To leverage this debiasing feature further, we propose a more general framework, \r2v, that can also compensate for other systematic biases in random walks.
We show that \r2v performs better than conventional embedding methods in link prediction and community detection tasks.
Using a citation graph of 260k journals, we demonstrate that the biases from random walks overshadow the salient features of graphs.
By removing the bias, \r2v better captures the characteristics of journals such as the impact factor and journal subject. The python code of residual2vec is available at GitHub~\cite{r2v}.

\section{Built-in debiasing feature of SGNS word2vec}
\label{sec:related_work}

\subsection{Background: SGNS word2vec}
Consider a sentence of words $(x_{1}, x_{2}, x_{3},\ldots)$ composed of $N$ unique words.
\w2v associates the $t$th word $x_{t}$ with words in its surrounding $x_{t-T}, \ldots, x_{t-1}, x_{t+1},\ldots, x_{t+T}$, which are referred to as context words, determined by a prescribed window size $T$.
For a center-context word pair $(i,j)$, \w2v models conditional probability
\begin{equation}
    P_{\text{w2v}}(j \vert i) = \frac{\exp(\vec{u}_i ^\top \vec{v}_{j})}{\sum_{j'=1}^N \exp(\vec{u}_{i}^\top \vec{v}_{j'})},\label{eq:cond_prob_w2v}
\end{equation}
where $\vec{u}_i, \vec{v}_i \in {\mathbb R}^{K\times 1}$ are \textit{embedding vectors} representing word $i$ as center and context words, respectively, and $K$ is the embedding dimension.
An approach to fit $P_{\text{w2v}}$ is the maximum likelihood estimation, which is computationally expensive because $P_{\text{w2v}}$ involves the sum over all words.
Alternatively, several heuristics have been proposed, among which \textit{negative sampling} is the most widely used~\cite{Mikolov2013,Perozzi2014,Grover2016}.

Negative sampling trains \w2v as follows.
Given a sentence, a center-context word pair $(i,j)$ is sampled and labeled as $Y_j=1$.
Additionally, one samples $k$ random word $\ell$ as candidate context words from a noise distribution $p_0(\ell)$, and then labels $(i,\ell)$ as $Y_{\ell} = 0$.
In general, a popular choice of the noise distribution $p_0(\ell)$ is based on word frequency, \ie $p_0(\ell) \propto P_\text{d} (\ell) ^\gamma$, where $P_{\text{d}} (\ell)$ is the fraction of word $\ell$ in the given sentence, and $\gamma$ is a hyper-parameter.
Negative sampling trains $\vec{u}_i$ and $\vec{v}_j$ such that its label $Y_j$ is well predicted by a logistic regression model
\begin{align}
    \label{eq:logistic-regress}
    P_{\text{NS}}(Y_{j} = 1; \vec{u}_{i}, \vec{v}_j) = \frac{1}{1 + \exp(-\vec{u}_i ^\top \vec{v}_{j})},
\end{align}
by maximizing its log-likelihood.

\subsection{Implicit debiasing by negative sampling}

Negative sampling efficiently produces a good representation~\cite{Mikolov2013}. An often overlooked fact is that negative sampling is a simplified version of Noise Contrastive Estimation (NCE)~\cite{Gutmann2010,Dyer2014}, and this simplification biases the model estimation. In the following, we show that this estimation bias gives rise to a built-in debiasing feature of SGNS word2vec.

\subsubsection*{Noise contrastive estimation}

NCE is a generic estimator for probability model $P_m$ of the form~\cite{Gutmann2010}:
\begin{align}
    \label{eq:nce-model}
    P_m(x) = \frac{ f(x;\theta) }{\sum_{ x' \in {\cal X}} f(x';\theta) },
\end{align}
where $f$ is a non-negative function of data $x$ in the set ${\cal X}$ of all possible values of $x$.
\w2v (Eq.~\eqref{eq:cond_prob_w2v}) is a special case of $P_m$, where $f(x) = \exp(x)$ and $x = u_i ^\top v_{j}$.
NCE estimates $P_m$ by solving the same task as negative sampling---classifying a positive example and $k$ randomly sampled negative examples using logistic regression---but based on a Bayesian framework~\cite{Gutmann2010,Dyer2014}.
Specifically, as prior knowledge, we know that $1$ in $1+k$ pairs are taken from the given data, which can be expressed as prior probabilities~\cite{Gutmann2010,Dyer2014}:
\begin{align}
    \label{eq:prior}
    P(Y_{j} = 1) = \frac{1}{k + 1},\quad P(Y_{j} = 0) = \frac{k}{k + 1}.
\end{align}
Assuming that the given data is generated from $P_m$, the positive example ($Y_j = 1$) and the negative examples ($Y_j = 0$) are sampled from $P_m$ and $p_0(j)$, respectively~\cite{Gutmann2010,Dyer2014}, i.e.,
\begin{align}
    \label{eq:class-cond}
    P(j \vert Y_{j} = 1) & = P_m(\vec{u}_i ^\top \vec{v}_{j}),
    \quad
    P(j \vert Y_{j} = 0)  = p_0 (j).
\end{align}
Substituting Eqs.~\eqref{eq:prior} and \eqref{eq:class-cond} into the Bayes rule yields the posterior probability for $Y_j$ given an example $j$~\cite{Gutmann2010,Dyer2014}:
\begin{align}
    \label{eq:posterior}
    P_{\text{NCE}}\left(Y_{j} =1 \vert j\right)  = \frac{
        P\left(j \vert Y_{j}=1\right)P(Y_{j}=1)
    }{
        \sum_{y \in \{0,1\}} P\left(j \vert Y_{j} = y\right)P(Y_{j} = y)
    }
    = \frac{
        P_m(\vec{u}_i ^\top \vec{v}_{j})
    }{
        P_m(\vec{u}_i ^\top \vec{v}_{j})  + kp_0(j)
    },
\end{align}
which can be rewritten with a sigmoid function as
\begin{align}
    \label{eq:nce}
    P_{\text{NCE}}\left(Y_{j}=1 \vert j\right)
     & = \frac{
        1
    }{
        1 + kp_0(j) / P_m(\vec{u}_i ^\top \vec{v}_{j})
    }
    = \frac{
        1
    }{
        1 + \exp\left[ - \ln f(\vec{u}_i ^\top \vec{v}_{j})  + \ln p_0(j) + c \right]
    },
\end{align}
where $c = \ln k + \ln\sum_{ x' \in {\cal X}} f(x')$ is a constant.
NCE learns $P_m$ by the logistic regression based on Eq.~\eqref{eq:nce}. The key feature of NCE is that it is an asymptomatically unbiased estimator of $P_m$ whose bias goes to zero as the number of training examples goes to infinity~\cite{Gutmann2010}.

\subsubsection*{Estimation bias of negative sampling}

In the original paper of word2vec~\cite{Mikolov2013}, the authors simplified NCE into negative sampling by dropping $\ln p_0(j) + c$ in Eq.~\eqref{eq:nce} because it reduced the computation and yielded a good word embedding. In the following, we show the impact of this simplification on the final embedding.

We rewrite $P_{\text{NS}}$ (\ie Eq.~\eqref{eq:cond_prob_w2v}) in the form of $P_{\text{NCE}}$ (\ie Eq.~\eqref{eq:nce}) as

\begin{align}
    P_{\text{NS}}(Y_{j} = 1; \vec{u}_{i}, \vec{v}_j) & = \frac{1}{1 + \exp(-\vec{u}_i ^\top \vec{v}_{j})}
    = \frac{
        1
    }{
        1 + \exp\left[ - \left( \vec{u}_i ^\top \vec{v}_{j} + \ln p_0(j) + c \right) + \ln p_0(j) + c \right]
    }                                             \nonumber                                               \\
                                                     & = \frac{
        1
    }{
        1 + \exp\left[ - \ln f(\vec{u}_i ^\top \vec{v}_{j}) + \ln p_0(j) + c \right]
    },
\end{align}
Equation (8) makes clear the relationship between negative sampling and NCE: negative sampling is the NCE with $f(\vec{u}_i ^\top \vec{v}_{j}) = \exp\left( \vec{u}_i ^\top \vec{v}_{j} + \ln p_0(j) + c\right)$ and noise distribution $p_0$~\cite{murray2020unsupervised}.
Bearing in mind that NCE is the asymptomatically unbiased estimator of Eq.~\eqref{eq:nce-model} and substituting $f(\vec{u}_i ^\top \vec{v}_{j})$ into Eq.~\eqref{eq:nce-model}, we show that SGNS word2vec is an asymptomatically unbiased estimator for probability model:
\begin{align}
    \label{eq:word2vecns}
    P_{\text{w2v}}^{\text{SGNS}}(j\given i) =  \frac{  p_0(j)\exp( \vec{u}_i ^\top \vec{v}_{j} ) }{Z_i '},\quad \text{where}\; Z_i ' := \sum_{j'=1}^N p_0(j') \exp(\vec{u}_{i} ^\top \vec{v}_{j'}).
\end{align}
Equation~\eqref{eq:word2vecns} clarifies the role of noise distribution $p_0$.
Noise probability $p_0$ serves as a baseline for $P_{\text{w2v}}^{\text{SGNS}}$, and word similarity $\vec{u}_i ^\top \vec{v}_{j}$ represents the \textit{deviation} from $p_0(j)$, or equivalently the characteristics of words \textit{not} captured in $p_0(j)$.
Notably, baseline $p_0(j) \propto P_\text{d}(j) ^\gamma$ is determined by word frequency $X_\text{d}(j)$ and thus negates the word frequency bias.
This realization---that we can explicitly use a noise distribution to obtain ``residual'' information---is the motivation for our method, \textit{residual2vec}.

\section{Residual2vec graph embedding}

We assume that the given graph is undirected and weighted, although our results can be generalized to directed graphs (see Supplementary Information). We allow multi-edges (i.e., multiple edges between the same node pair) and self-loops, and consider unweighted graphs as weighted graphs with all edge weight set to one~\cite{Fortunato2010,Newman2014}.

\subsection{Model}

\begin{figure}
    \centering
    \includegraphics[width=\hsize]{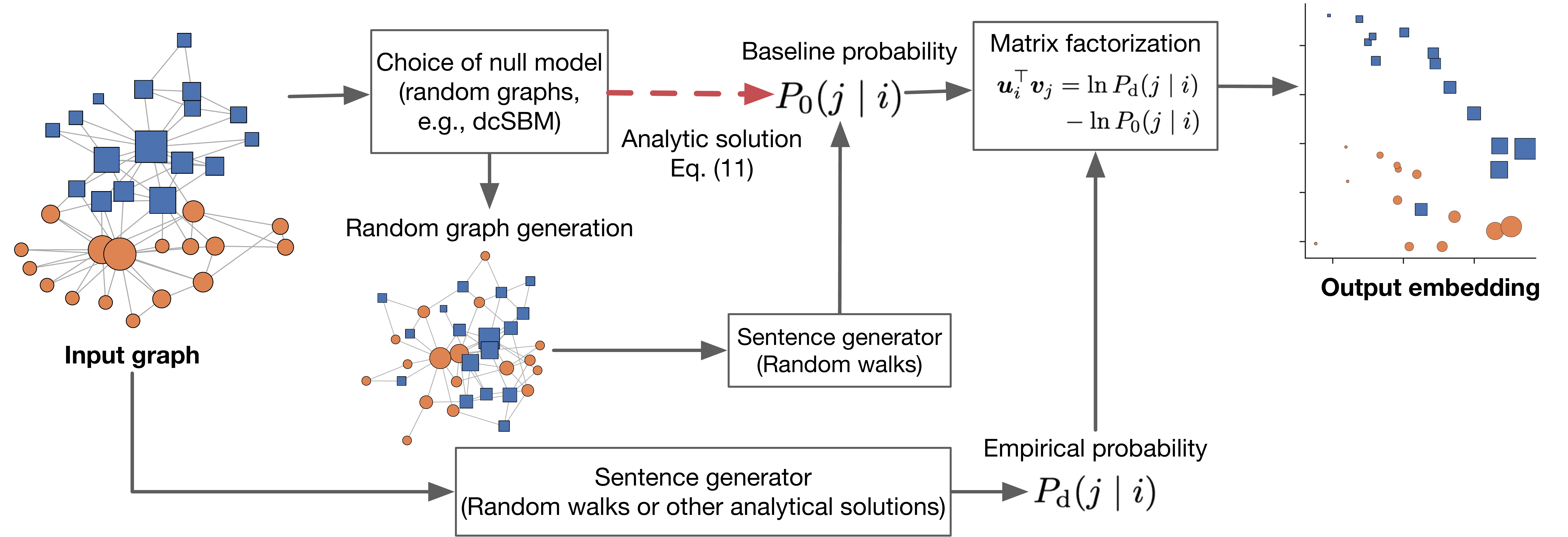}
    \caption{
        \r2v framework.
        We first choose an explicit ``null'' model (\eg dcSBM random graph).
        By running random walks in the input and random graphs, ``sentences of nodes'' are generated and then
        compared to extract the residual information not in the random graphs.
        The final embedding encodes the residual information.
        Probabilities $P_{\text{d}}(j\given i)$ and $P_{0}(j\given i)$ can be computed analytically in many cases.
        For instance, we use the analytical solution for dcSBM (\ie $P_{\text{d}}(j \given i)$ and $P_0(j \given i)$) instead of simulating actual random walks.
    }
    \label{fig:r2v-schematic}
\end{figure}

The presence of $p_0(j)$ effectively negates the bias in random walks due to degree.
This bias dictates that, for a sufficiently long trajectory of random walks in undirected graphs, the frequency $P_\text{d}(j)$ of node $j$ is proportional to degree $d_j$ (\ie the number of neighbors) irrespective of the graph structure~\cite{MASUDA2017b}.
Now, if we set $\gamma=1$, baseline $p_0(j)=P_\text{d}(j)$ matches exactly with the node frequency in the trajectory, negating the bias due to degree.
But, we are free to choose any $p_0$.
This consideration leads us to \textit{residual2vec} model:
\begin{align}
    \label{eq:model_residual2vec}
    P_{\text{r2v}}(j\given i) =  \frac{P_{0}(j \given i)\exp( \vec{u}_i ^\top \vec{v}_{j} )}{Z_i'},
\end{align}
where we explicitly model baseline transition probability denoted by $P_{0}(j \given i)$.
In doing so, we can obtain the \textit{residual} information that is not captured in $P_{0}$.
Figure~\ref{fig:r2v-schematic} shows the framework of \r2v.
To negate a bias, we consider ``null'' graphs, where edges are randomized while keeping the property inducing the bias intact~\cite{Fortunato2010,Newman2014}.
Then, we compute $P_{0}$ either analytically or by running random walks in the null graphs.
The $P_0$ is then used as the noise distribution to train SGNS \w2v.

\subsubsection*{Random graph models}
Among many models for random graph~\cite{erdHos1959random,Karrer2009,Garlaschelli2009,Karrer2011,Expert2011,Levy2014,fosdick2018}, here we focus on the degree-corrected stochastic block model (dcSBM), which can be reduced to many fundamental random graph models with certain parameter choices~\cite{Karrer2011}.
With the dcSBM, one partitions nodes into $B$ groups and randomizes edges while preserving (i) the degree of each node, and (ii) the number of inter-/intra-group edges.
Preserving such group connectivity is useful to negate biases arising from less relevant group structure such as bipartite and multilayer structures.
The dcSBM can be mapped to many canonical ensembles that preserve the expectation of structural properties. In fact, when $B=1$, the dcSBM is reduced to the soft configuration model that preserves the degree of each node on average, with self-loops and multi-edges allowed~\cite{fosdick2018}. Furthermore, by setting $B = 1$ and $d_i=\text{constant}$, the dcSBM is reduced to the Erdős-Rényi model for multigraphs that preserves the number of edges on average, with self-loops and multi-edges allowed.
In the dcSBM, the edge weights follow a Poisson distribution and thus take integer values.

Suppose that the nodes in the given graph have $B$ discrete labels (\eg gender), and we want to remove the structural bias associated with the labels.
If no such label is available, all nodes are considered to have the same label (\ie $B=1$).
We fit the dcSBM with $B$ groups, where each group consists of the nodes with the same label. The dcSBM generates random graphs that preserve the number of edges within and between the groups (\eg assortativity by gender types).
We can calculate $P_{0}(j\given i)$ without explicitly generating random graphs (Supplementary Information)
\begin{align}
    \label{eq:pnull}
    P_{0}(j \given i) =\frac{d_j}{D_{g_j}} \left( \frac{1}{T}\sum_{t=1}^T \mat{P}_{\text{SBM}} ^t \right)_{g_i, g_j},
\end{align}
where node $j$ has degree $d_j$ and belongs to group $g_j$, $D_{g}=\sum_{\ell=1}^N d_{\ell} \delta(g_\ell, g)$, and $\delta$ is Kronecker delta.
The entry $P^{\text{SBM}} _{g,g'}$ of matrix $\mat{P}_{\text{SBM}} = (P^{\text{SBM}} _{g,g'}) \in \mathbb{R}^{B\times B}$ is the fraction of edges to group $g'$ in $D_g$.
Table~\ref{ta:null_models} lists $P_{0}(j\given i)$ for the special classes of the dcSBM.
See Supplementary Information for the step-by-step derivation.

\begin{table}
    \centering
    \caption{
        Baseline probability $P_{0}(j \given i)$ for the dcSBM along with its special cases.
    }
    \label{ta:null_models}
    \scalebox{0.9}{
        \begin{tabular}{lp{10em}p{12em}p{12em}} \toprule
                                         & \multicolumn{3}{c}{Canonical random graph models}                                                                                                                                                                                                     \\\cmidrule{2-4}
                                         & \ErdosRenyi model for multigraphs~\cite{erdHos1959random}                                          & \multirow{1}{10em}{The soft configuration model~\cite{fosdick2018}} & \multirow{2}{10em}{dcSBM~\cite{Karrer2011}}                                                                                   \\ \midrule
            $P_0(j \given i)$            & \(\displaystyle 1/N \)                                                             & $d_j / \sum_{\ell=1}^N d_{\ell}$           & \(\displaystyle d_j D_{g_j} ^{-1}\left( T^{-1}\sum_{t=1}^T \mat{P}_{\text{SBM}} ^t \right)_{g_i, g_j}  \) \\
            \multirow{2}{5em}{Algorithm}                    &
            \multirow{2}{5em}{\DeepWalk~\cite{Perozzi2014}} & \n2v ($\gamma=0.75$)~\cite{Grover2016} &                                                                                                                                                        \\ \
            &
            & \method{NetMF} ($\gamma=1$)~\cite{Qiu2018} &                                                                                                                                                        \\ \bottomrule
        \end{tabular}
    }
\end{table}

\subsubsection*{Other graph embedding as special cases of residual2vec}%
\r2v can be considered as a general framework to understand structural graph embedding methods because many existing graph embedding methods are special cases of \r2v.
\n2v and \method{NetMF} use SGNS \w2v with $p_0(j) \propto P_\text{d}(j) ^\gamma$.
This $p_0(j)$ is equivalent to the baseline for the soft configuration model~\cite{fosdick2018}, where each node $i$ has degree $d_i^\gamma$.
\DeepWalk is also based on \w2v but trained with an unbiased estimator (\ie the hierarchical softmax).
Because negative sampling with $p_0(j) = 1/N$ is unbiased~\cite{Gutmann2010,Dyer2014}, \DeepWalk is equivalent to \r2v with the \ErdosRenyi random graphs for multigraphs.


%

\subsection{Residual2vec as matrix factorization} %
Many structural graph embedding methods implicitly factorize a matrix to find embeddings~\cite{Qiu2018}.
\r2v can also be described as factorizing a matrix $\mat{R}$ which captures residual pointwise mutual information.
Just like a $K$th order polynomial function can be fit to $K$ points without errors, we can fit \w2v to the given data without errors when the embedding dimension $K$ is equal to the number of unique words $N$~\cite{Levy2014}.
In other words, $P_{\text{d}}(j \given i)=P_{\text{r2v}}(j \given i),\; \forall i,j$ if $K=N$.
By substituting $P_{\text{r2v}}(j \given i) = P_0(j \given i)\exp(\vec{u}^\top _i \vec{v}_j)/Z' _i$, we obtain
\begin{align}
    P_{\text{d}}(j \given i)
     &
    = \frac{P_{\text{d}}(j \given i)\exp\left( \vec{u}_i ^\top \vec{v}_{j} + \ln P_{0}(j \given i)-\ln P_{\text{d}}(j \given i)\right) }{\sum_{j'=1}^N P_{\text{d}}(j' \given i)\exp\left( \vec{u}_i ^\top \vec{v}_{j'} + \ln P_{0}(j' \given i) - \ln P_{\text{d}}(j' \given i)\right)}.
\end{align}
The equality holds if $\exp( \vec{u}_i ^\top \vec{v}_{j} + \ln P_{0}(j \given i )-\ln P_{\text{d}}(j \given i)) = c_i$ for all $i$ and $j$, where $c_i$ is a constant.
The solution is not unique because $c_i$ can be any real value.
We choose $c_i=1$ to obtain a solution in the simplest form, yielding matrix $\mat{R}$ that \r2v factorizes:
\begin{align}
    \label{eq:root_finding}
    R_{ij}:= \vec{u}_i ^\top \vec{v}_j = \ln P_{\text{d}}(j \given i) - \ln P_{0}(j\given i).
\end{align}
Matrix $\mat{R}$ has an information-theoretic interpretation. We rewrite
\begin{align}
    \label{eq:root_finding2}
    R_{ij} & = \ln \frac{P_{\text{d}}(i,j)}{P_\text{d}(i)}-\ln \frac{P_{\text{0}}(i,j)}{P_\text{0}(i)}
    = \ln \frac{P_{\text{d}}(i,j)}{P_\text{d}(i)P_\text{d}(j)}-\ln \frac{P_{\text{0}}(i,j)}{P_\text{0}(i)P_\text{0}(j)} + \ln P_\text{d}(j) - \ln P_\text{0}(j).
\end{align}
The dcSBM preserves the degree of each node and thus has the same the degree bias with the given graph, \ie $P_\text{d}(i) = P_\text{0}(i)$ (Supplementary Information), which leads
\begin{align}
    \label{eq:root_finding3}
    R_{ij} & =  \text{PMI}_{\sim P_\text{d}}(i,j) - \text{PMI}_{\sim P_\text{0}}(i,j),\;\text{where}\; \text{PMI}_{\sim P}(i,j) = \ln \frac{P(i,j)}{P(i)P(j)}.
\end{align}
$\text{PMI}_{\sim P}(i,j)$ is the pointwise mutual information that measures the correlation between center $i$ and context $j$ under joint distribution $P(i,j)$, \ie $\text{PMI}_{\sim P}(i,j) = 0$ if
$i$ and $j$ appear independently, and $\text{PMI}_{\sim P}(i,j) > 0$ otherwise.
In sum, $R_{ij}$ reflects residual pointwise mutual information of $i$ and $j$ from the null model.

\subsection{Efficient matrix factorization}

Although we assume that $N=K$ above, in practice, we want to find a compact vector representation (\ie $K\ll N$) that
still yields a good approximation~\cite{Levy2014,Qiu2018}.
There are several computational challenges in factorizing $\mat{R}$.
First, $\mat{R}$ is ill-defined for any node pair $(i,j)$ that never appears because $R_{ij}=\ln 0 =-\infty$.
Second, $\mat{R}$ is often a dense matrix with ${\cal O}(N^2)$ space complexity.
For these issues, a common remedy is a truncation~\cite{Levy2014,Qiu2018}:
\begin{align}
    \tilde R_{ij}:= \max(R_{ij}, 0).
\end{align}
This truncation discards negative node associations ($R_{ij}<0$) while keeping the positive associations ($R_{ij}>0$) based on the idea that negative associations are common, and thus are less informative~\cite{Qiu2018}.
In both word and graph embeddings, the truncation substantially reduces the computation cost of the matrix factorization~\cite{Levy2014,Qiu2018}.

We factorize $\mat{\tilde R}=(\tilde R_{ij})$ into embedding vectors $\vec{u}_i$ and $\vec{v}_j$ such that $\vec{u}_i ^\top \vec{v}_j \simeq \tilde R_{ij}$ by
using the truncated singular value decomposition (SVD).
Specifically,
we factorize $\mat{\tilde R}$ by $\mat{\tilde R} = \mat{\Phi}_{\text{left}} \cdot \diag(\sigma_1, \sigma_2, \ldots, \sigma_K)\cdot \mat{\Phi}_{\text{right}} ^\top$,
where $\mat{\Phi}_{\text{left}}=(\phi_{ik} ^{\text{left}})_{ik} \in {\mathbb R}^{N\times K}$ and $\mat{\Phi}_{\text{right}}=(\phi^{\text{right}}_{ik})_{ik} \in {\mathbb R}^{N\times K}$ are the left and right singular vectors of $\mat{\tilde R}$ associated with the $K$ largest singular values ($\sigma_1,\ldots \sigma_K$) in magnitude, respectively.
Then, we compute $\vec{u}_i$ and $\vec{v}_j$ by
$u_{ik} = \sigma_k^{\alpha}\phi_{ik}^{\text{left}}, \; v_{ik} = \sigma_k^{1-\alpha}\phi_{ik}^{\text{right}}$ with $\alpha = 0.5$ following the previous studies~\cite{Levy2014,Qiu2018}.

The analytical computation of $P_{\text{d}}(j \given i)$ is expensive because it scales as ${\cal O}(TN^3)$, where $T$ is the window size~\cite{Qiu2018}.
Alternatively, one can simulate random walks to estimate $P_{\text{d}}(j \given i)$.
Yet, both approaches require ${\cal O}(N^2)$ space complexity.
Here, we reduce the time and space complexity by \textit{the block approximation} that approximates the given graph by the dcSBM with $\hat B$ groups (we set $\hat B=1,000$) and then computes an approximated $P_{\text{d}}(j \given i)$.
The block approximation reduces the time and space complexity to ${\cal O}((N + M)\hat B + T\hat B^3)$ and ${\cal O}(NB\hat B)$ for a graph with $N$ nodes and $M$ edges, respectively, with a high accuracy (the average Pearson correlation of 0.85 for the graphs in Table~\ref{ta:network_list}).
See Supplementary Information for the block approximation.

\begin{table}
    \centering
    \caption{
        List of empirical graphs and structural properties. Variables $N$ and $M$ indicate the number of nodes and edges, respectively.
    }
    \scalebox{0.75}{
        \begin{tabular}{lllllll}
            \toprule
            Network                & $N$    & $M$     & Assortativity & Max. degree & Clustering coef. & Ref.                \\
            \midrule
            Airport                & 3,188  & 18,834  & -0.02         & 911         & 0.493            & \cite{ToreOpsahl}   \\
            Protein-Protein        & 3,852  & 37,840  & -0.10         & 593         & 0.148            & \cite{Stark2006}    \\
            Wikipedia vote         & 7,066  & 100,736 & -0.08         & 1065        & 0.142            & \cite{Leskovec2010} \\
            Coauthorship (HepTh)   & 8,638  & 24,827  & 0.24          & 65          & 0.482            & \cite{Leskovec2007} \\
            Citation (DBLP)        & 12,494 & 49,594  & -0.05         & 713         & 0.118            & \cite{Ley2002}      \\
            Coauthorship (AstroPh) & 17,903 & 197,031 & 0.20          & 504         & 0.633            & \cite{Leskovec2007} \\
            \bottomrule
        \end{tabular}
    }
    \label{ta:network_list}
\end{table}

\section{Results}
\label{sec:results}

We test \r2v using link prediction and community detection benchmarks~\cite{Lancichinetti2009,Newman2006a,Fortunato2010,Grover2016,abu2017learning}.
We use the soft configuration model~\cite{fosdick2018} as the null graph for \r2v, denoted by \method{r2v-config}, which yields a degree-debiased embedding.
The soft configuration model allows self-loops and multi-edges---which are not present in the graphs used in the benchmarks---and thus is not perfectly compatible with the benchmark graphs. Nevertheless, because the multi-edges and self-loops are rare in the case of sparse graphs, the soft configuration model has been widely used for sparse graphs without multi-edges and self-loops~\cite{Newman2006a,Newman2014,fosdick2018}.

As baselines, we use (i) three random-walk-based methods, \n2v~\cite{Grover2016}, \DeepWalk~\cite{Perozzi2014}, and \FairWalk~\cite{Rahman2019-uh},
(ii) two matrix-factorization-based methods, \Glove~\cite{Pennington2014} and Laplacian eigenmap (\LEM)~\cite{Belkin2003},
and (iii) the graph convolutional network (\GCN)~\cite{kipf2017semi}, the graph attention networks (\GAT)~\cite{Velickovic2018-tl}, and \GraphSAGE~\cite{Hamilton2017}.
For all random-walk-based methods, we run 10 walkers per node for 80 steps and set $T=10$ and training iterations to 5.
We set the parameters of \n2v by $p=1$ and $q \in \{0.5, 1, 2\}$.
For \Glove, we input the sentences generated by random walks.
We use two-layer \GCN, \GraphSAGE, and \GAT implemented in StellarGraph package~\cite{StellarGraph} with the parameter sets (\eg the number of layers and activation function) used in \cite{kipf2017semi,Hamilton2017,Velickovic2018-tl}.
Because node features are not available in the benchmarks, we alternatively use degree and eigenvectors for the $\hat K$ smallest eigenvalues of the normalized Laplacian matrix because they are useful for link prediction and clustering~\cite{Luxburg2007,Kunegis2009}.
We set $\hat K$ to dimension $K$ (\ie $\hat K = K$). Increasing $\hat K$ does not improve performance much (Supplementary Information).

\begin{figure}
    \centering
    \includegraphics[width=\textwidth]{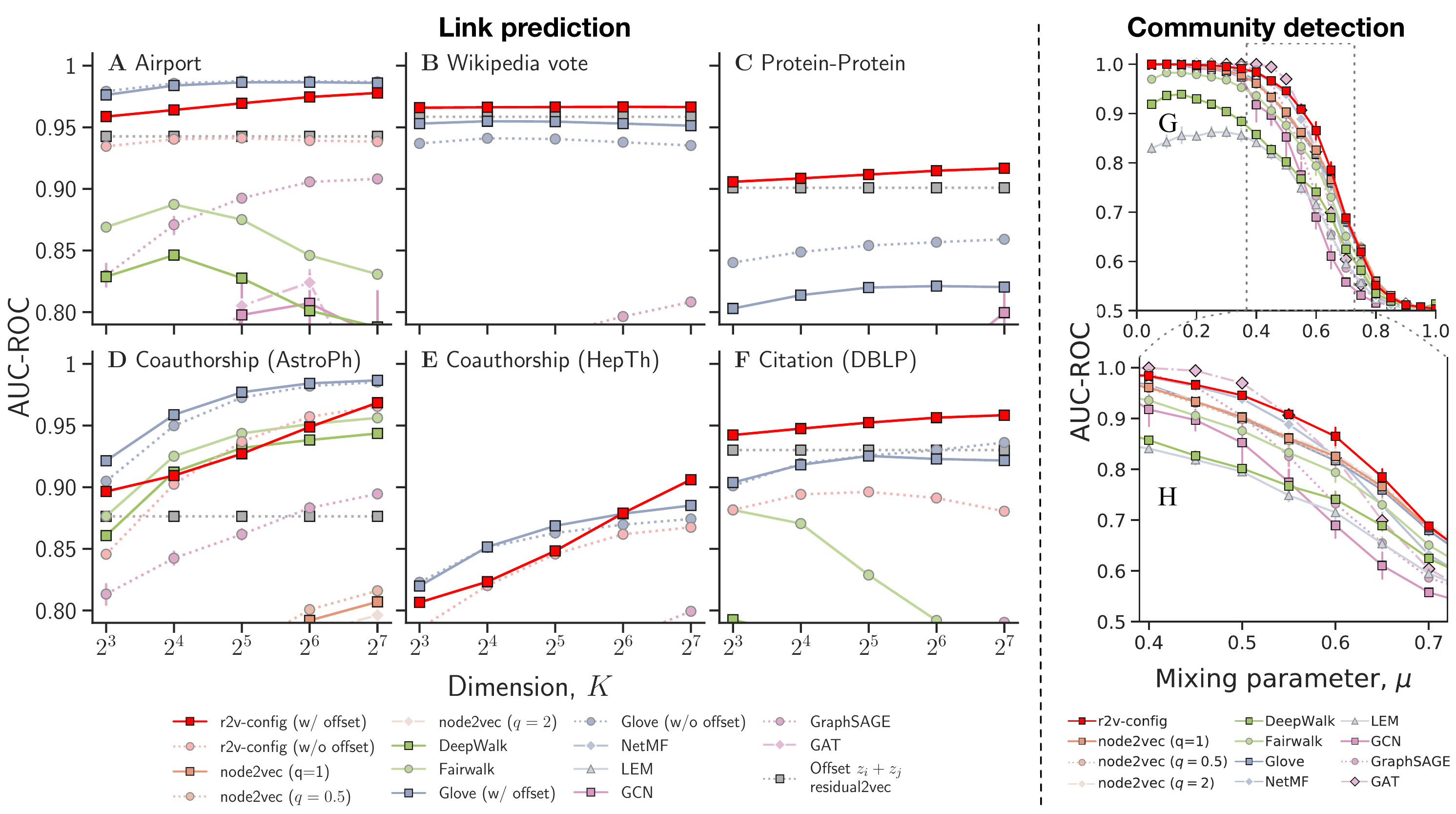}
    \caption{
        Performance for link prediction and community detection.
        We average AUC-ROC values over 30 runs with different random seeds and compute
        the 90\% confidence interval by a bootstrapping with $10^4$ resamples.
        For link prediction, some underperforming models are not shown.
        See Supplemantary Information for the full results.
        \method{r2v-config} consistently performs the best or nearly the best for all graphs in both benchmarks, as indicated by the higher AUC-ROC values.
    }
    \label{fig:benchmarks}
\end{figure}

\subsubsection*{Link prediction}
Link prediction task is to find missing edges based on graph structure, a basic task for various applications such as recommending friends and products~\cite{Grover2016,abu2017learning,zhang2018arbitrary}.
The link prediction task consists of the following three steps.
First, given a graph, a fraction ($\rho=0.5$) of edges are randomly removed.
Second, the edge-removed graph is embedded using a graph embedding method.
Third, the removed edges are predicted based on a likelihood score calculated based on the generated embedding.
In the edge removal process, we keep edges in a minimum spanning tree of the graph to ensure that the graph is a connected component~\cite{Grover2016,abu2017learning}.
This is because predicting edges between disconnected graphs is an ill-defined task because each disconnected component has no relation to the other.

We leverage both embedding $\vec{u}_i$ and baseline probability $P_0(j \vert i)$ to predict missing edges. Specifically, we calculate the prediction score by $\vec{u}_i ^\top \vec{u}_j + z_i + z_j$, where we set $z_j=\ln P_0(j \vert i)$ for residual2vec because $\ln P_0(j \vert i)$ has the same unit as $\vec{u}_i ^\top \vec{u}_j$ (Supplementary Information). Glove has a bias term that is equivalent to $z_i$. Therefore, we set $z_i$ to the bias term for Glove. Other methods do not have the parameter that corresponds to $z_i$ and thus we set $z_i = 0$.
We measure the performance by the area under the curve of the receiver operating characteristics (AUC-ROC) for the prediction scores, with the removed edges and the same number of randomly sampled non-existent edges being the positive and negative classes, respectively.
We perform the benchmark for the graphs in Table~\ref{ta:network_list}.

\method{r2v-config} performs the best or nearly the best for all graphs (Figs.~\ref{fig:benchmarks}A--F).
It consistently outperforms other random walk-based methods in all cases despite the fact that \n2v and \method{r2v-config} train the same model.
The two methods have two key differences.
First, \method{r2v-config} uses baseline $P_0(\ell \given i) = P_{\text{d}}(\ell)$, whereas \n2v uses $P_0(\ell \given i) \propto P_{\text{d}}(\ell)^{3/4}$ that does not exactly fit to the degree bias.
Second, \method{r2v-config} optimizes the model based on a matrix factorization, which often yields a better embedding than the stochastic gradient descent algorithm used in \n2v~\cite{Levy2014,Qiu2018}.
The performance of \r2v is substantially improved when incorporating offset $z_i$, which itself is a strong predictor as indicated by the high AUC-ROC.


\subsubsection*{Community detection}
We use the Lancichinetti–Fortunato–Radicchi (LFR) community detection benchmark~\cite{Lancichinetti2009}.
The LFR benchmark generates graphs having groups of densely connected nodes (\ie communities) with a power-law degree distribution with a prescribed exponent $\tau$.
We set $\tau = 3$ to generate the degree heterogeneous graphs. See Supplementary Information for the case of degree homogeneous graphs.
In the LFR benchmark, each node has, on average, a specified fraction $\mu$ of neighbors in different communities.
We generate graphs of $N=1,000$ nodes with the parameters used in Ref.~\cite{Lancichinetti2009} and embed the graphs to $K=64$ dimensional space.
We evaluate the performance by randomly sampling $10,000$ node pairs and calculate the AUC-ROC for their cosine similarities, with nodes in the same and different communities being the positive and negative classes, respectively.
A large AUC value indicates that nodes in the same community tend to have a higher similarity than those in different communities.

As $\mu$ increases from zero, the AUC for all methods decreases because nodes have more neighbors in different communities.
\DeepWalk and \LEM have a small AUC value even at $\mu=0.05$.
\method{r2v-config} consistently achieves the highest or the second-highest AUC.

\subsubsection*{Case study}

\begin{figure}
    \centering
    \includegraphics[width=0.95\hsize]{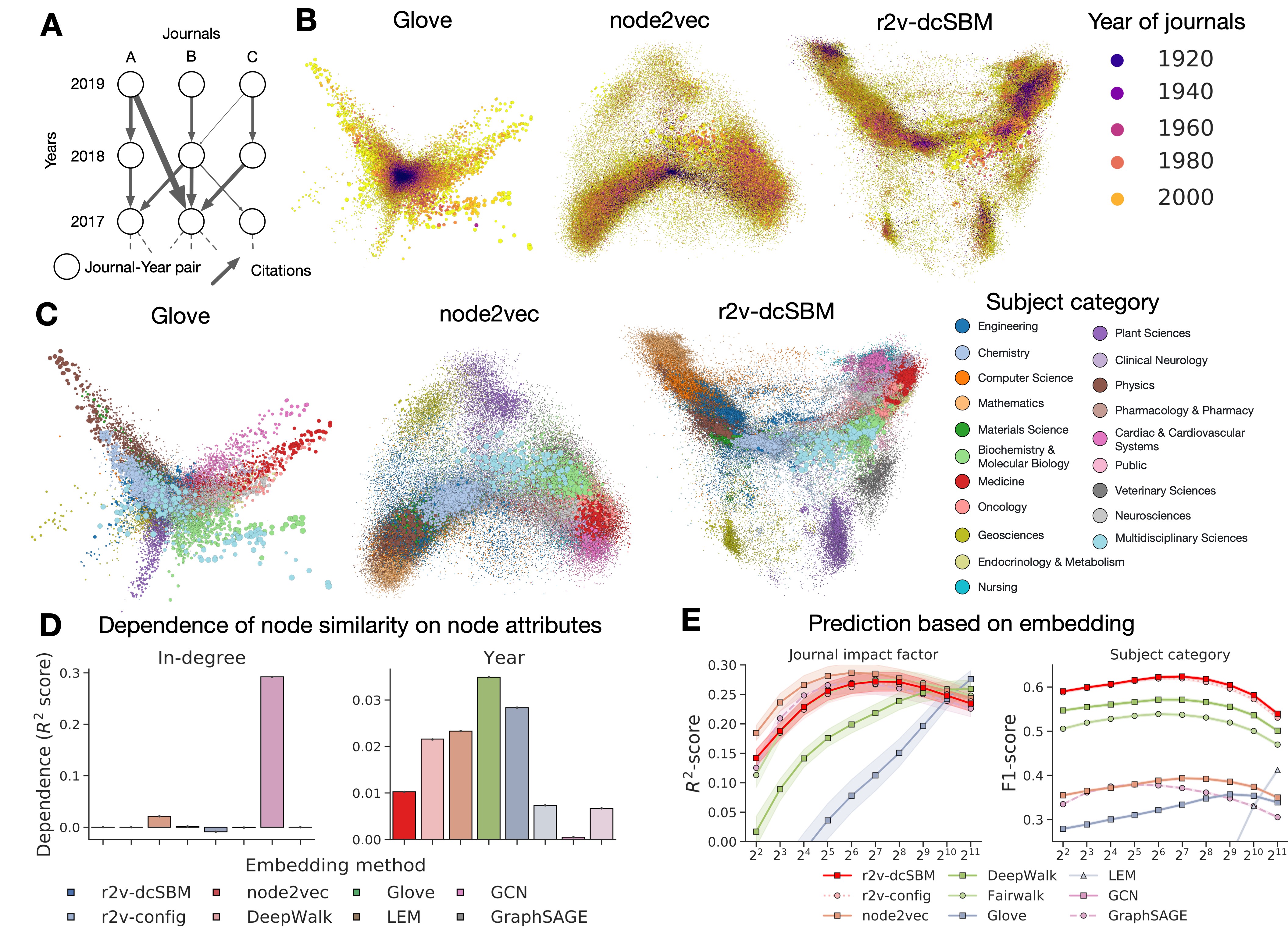}
    \caption{
        Embedding of the WOS journal citation graph.
        ({\bf A}) Each node represents a pair $(j,t)$ of journal $j$ and year $t$, and each edge is weighted by citations.
        ({\bf B}, {\bf C}) A 2d projection of 128-dimensional embedding by the LDA.
        ({\bf D}) \method{r2v-dcSBM} produces the embedding that is less dependent on degree and time.
        ({\bf E}) By using the k-nearest neighbor algorithm, the embedding by \method{r2v-dcSBM} best predicts
        the impact factor and subject category.
    }
    \label{fig:wos-journal-embedding}
\end{figure}

Can debiasing reveal the salient structure of graphs more clearly?
We construct a journal citation graph using citation data between 1900 and 2019 indexed in the Web of Science (WoS)  (Fig.~\ref{fig:wos-journal-embedding}A).
Each node represents a pair $(i, t)$ of journal $i$ and year $t$.
Each undirected edge between $(i,t)$ and $(j,t')$ is weighted by the number of citations between journal $i$ in year $t$ and journal $j$ in year $t'$.
The graph consists of $242,789$ nodes and $254,793,567$ undirected and weighted edges.
Because the graph has a high average degree (\ie $1,049$), some algorithms are computationally demanding.
For this reason, we omit \n2v with $q=0.5$ and $q=2$, \method{NetMF}, and \GAT due to memory shortage (1Tb of RAM).
Furthermore, for \GCN, we set $\hat K=32$, which still took more than 18 hours.
We also perform the whole analysis for the directed graph to respect the directionality of citations.
Although all methods perform worse in predicting impact factor and subject category, we find qualitatively the same results.
See Supplementary Information for the results for the directed graph.

Here, in addition to the degree bias, there are also temporal biases, \eg there has been an exponential growth in publications, older papers had more time to accumulate citations, and
papers tend to cite those published in prior few years~\cite{Wang2013}.
To remove both biases, we use \r2v with the dcSBM (denoted by \method{r2v-dcSBM}), where we group journals by year to randomize edges while preserving the number of citations within and between years.
We generate $K=128$ dimensional embeddings with $T=10$.

Figures~\ref{fig:wos-journal-embedding}B and C show the 2d projection by the Linear Discriminant Analysis (LDA) with journals' subject categories as the class labels.
\Glove and \n2v capture the temporal structure prominently, placing many old issues at the center of the embeddings. By contrast, \method{r2v-dcSBM} spreads out the old issues on the embedding.
To quantify the effect of temporal bias, we randomly sample $50,000$ node pairs $(i,j)$ and then fitting a linear regression model $y_{ij} = w_{0} + w_1 (x_i + x_j) + w_2 |x_i - x_j| + w_3 x_i x_j$ that predicts cosine similarity $y_{ij}$ for node pair $(i,j)$ with attributes $x_i$ and $x_j$, where $x_i$ is either the degree or the year of node $i$.
We perform 5-cross validations and compute $R^2$-score (Fig.~\ref{fig:wos-journal-embedding}D).
A smaller $R^2$-score indicates that node similarity is less dependent on node attributes and thus less biased.
\LEM has the smallest $R^2$-score for both degree and year.
\method{r2v-dcSBM} has a smaller $R^2$-score than \method{r2v-config} for year, respectively, suggesting that
\method{r2v-dcSBM} successfully negates the biases due to time.

Is debiasing useful to capture the more relevant structure of graphs?
We use embedding vectors to predict journal's impact factor (IF) and subject category.
By employing the $k$-nearest neighbor algorithm, we carry out $5$-cross validations and measure the prediction performance by $R^2$-score and the micro-F1 score.
To ensure that the train and test sets do not have the same journals in the cross-validation, we split the set of journals $i$ into the train and test sets instead of splitting the set of nodes ($i$,$t$).
No single method best predicts both impact and subject categories. Yet, \method{r2v-config} and \method{r2v-dcSBM} consistently achieve the strongest or nearly the strongest prediction power for all $k$ we tested (Fig.~\ref{fig:wos-journal-embedding}E).
This result demonstrates that debiasing embedding can reveal the salient structure of graphs that is overshadowed by other systematic biases.

\section{Discussion}
\label{sec:discussion}

In this paper, starting from the insight that \w2v with SGNS has a built-in debiasing feature that cancels out the bias due to the degree of nodes,
we generalize this debiasing feature further, proposing a method that can selectively remove any structural biases that are modeled by a null random graph.
By exposing the bias and explicitly modeling it, we provide a new way to integrate prior knowledge about graphs into graph embedding, and a unifying framework to understand structural graph embedding methods.
Under our residual2vec framework, other structural graph embedding methods that use random walks can be understood as special cases with different choices of null models.
Through empirical evaluations, we demonstrate that debiasing improves link prediction and community detection performances, and
better reveals the characteristics of nodes, as exemplified in the embedding of the WoS journal citation graph.

Our method is highly flexible because any random graph model can be used.
Although we focus on two biases arising from degree and group structure in a graph, one can remove other biases such as
the degree-degree correlation, clustering, and bipartitivity by considering appropriate null graphs.
Beyond these statistical biases, there have been growing concerns about social bias (\eg gender stereotype) as well as surveillance and privacy in AI applications, which
prompted the study of gender and frequency biases in word embedding~\cite{Bolukbasi2016,ethayarajh-etal-2019-towards,ethayarajh-etal-2019-understanding,Zhou2021}.
The flexibility and power of our selective and explicit debiasing approach may also be useful to address such biases that do not originate from common graph structures.

There are several limitations in \r2v.
We assume that random walks have a stationary distribution, which may not be the case for directed graphs.
One can ensure the stationarity in random walks by randomly teleporting walkers~\cite{Lambiotte2012}.
Second, it is not yet clear to what extent debiasing affects downstream tasks (\eg by losing information about the original graph).
Nevertheless, we believe that the ability to understand and control systematic biases is critical to model graphs through the prism of embedding.

\section*{Broader Impact}
\label{sec:broad_impact}

There has been an ever-increasing concern on inappropriate social stereotyping and the leak of sensitive information in word and graph embeddings~\cite{Bolukbasi2016,Edwards2016}.
Although we have not studied social biases in this paper, given the wide usage of graph embedding methods to model social data, our approach may lead to methods and studies that expose and mitigate social biases that manifest as structural properties in graph datasets.
Our general idea and approach may also be applied to modeling natural language and may contribute to the study of biases in language models.
At the same time, by improving the accuracy of graph embedding, our method may also have negative impacts such as privacy attacks and exploitation of personal data (surveillance capitalism)~\cite{Bose2019,Edwards2016}.
Nevertheless, we believe that our approach contributes to the effort to create transparent and accountable machine learning methods, especially because our method enables us to explicitly model what is structurally expected.

\section*{Disclosure of Funding}
The authors acknowledge support from the Air Force Office of Scientific Research under award number FA9550-19-1-0391.

\small



\section*{Supplementary material}

# Supplementary Information: Residual2Vec: Debiasing graph embedding with random graphs

Kojaku et al.

# Contents

# 1 Supplementary information for derivation

## 1.1 $P_{\mathrm{d}}(j \mid i)$ for the given graph

We compute conditional probability $P(j \mid i)$ for the given and random graphs to construct matrix $\mathbf{R}$ to factorize. While many methods simulate random walks, we can analytically compute $P(j \mid i)$ by exploiting the properties of the random walks. Denoted by $(x_1, x_2, x_3, \ldots)$ the sentence generated by a random walk in an undirected graph. For center-context node pair $(i, j)$, $P(j \mid i)$ is given by

$$P_{\mathrm{d}}(j \mid i) = \frac{1}{2T} \sum_{\tau=1}^{T} P_{\mathrm{rw}}(x_{t-\tau} = j \mid x_t = i) + \frac{1}{2T} \sum_{\tau=1}^{T} P_{\mathrm{rw}}(x_{t+\tau} = j \mid x_t = i), \quad (1)$$

where $P_{\mathrm{rw}}(x_{t'} = j \mid x_t = i)$ is the probability that the walker visits $j$ at time $t'$ given that it visits $i$ at time $t$. The first and second terms of the right-hand side of Eq. (1) represent the probabilities of visiting $j$ *before* and *after* visiting $i$, respectively. Both terms are equal for an undirected and connected graph thanks to the following two key properties of the random walks. First, the random walk has a time-invariant stationary distribution, i.e., $P_{\mathrm{rw}}(x_{t-\tau} = j) = P_{\mathrm{rw}}(x_t = j) = P_{\mathrm{rw}}(j)$, where $P_{\mathrm{rw}}(j)$ is the stationary distribution at node $j$ [10]. Second, the random walks satisfy the detailed balanced condition [10], i.e., $P_{\mathrm{rw}}(i) P_{\mathrm{rw}}(x_t = j \mid x_{t-\tau} = i) = P_{\mathrm{rw}}(j) P_{\mathrm{rw}}(x_t = i \mid x_{t-\tau} = j)$. With these properties as well as the chain rule of probability, the first term can be rewritten as the second term:

$$P_{\mathrm{rw}}(x_{t-\tau} = j \mid x_t = i) = \frac{P_{\mathrm{rw}}(j) P_{\mathrm{rw}}(x_t = i \mid x_{t-\tau} = j)}{P_{\mathrm{rw}}(i)} = P_{\mathrm{rw}}(x_t = j \mid x_{t-\tau} = i). \quad (2)$$

Returning to Eq. (1), conditional probability $P(j \mid i)$ can be computed analytically by

$$P_{\mathrm{d}}(j \mid i) = \frac{1}{T} \sum_{\tau=1}^{T} P_{\mathrm{rw}}(x_{t+\tau} = j \mid x_t = i) = \left( \frac{1}{T} \sum_{t=1}^{T} \mathbf{P}^t \right)_{ij}, \quad (3)$$

where $\mathbf{P} = (P_{ij})$ is the *transition matrix* of random walks in the graph, with entry $P_{ij}$ indicating the probability of moving from $i$ to $j$ by one step.

## 1.2 $P_{\mathbf{0}}(j \mid i)$ for the dcSBM

In the dcSBM, each edge between nodes $i$ and $j$ appears independently and has a weight $w_{ij}$ following a Poisson distribution with mean

$$\langle \tilde{w}_{ij} \rangle = \lambda_{g_i, g_j} \theta_i \theta_j, \quad (4)$$

where $\langle \cdot \rangle$ is the expectation over the Poisson distribution, and $\{g_i\}_i$, $\{\lambda_{g,g'}\}_{g,g'}$, and $\{\theta_i\}_i$ are the parameters of the dcSBM. Parameter $g_i$ indicates the group to which node $i$ belongs, and $\lambda_{g,g'}$ represents the number of edges between groups $g$ and $g'$. Parameter $\theta_i$ is a propensity of node $i$ normalized such that

$$\sum_{i=1, g_i=g}^{N} \theta_i = 1, \ \forall 1 \leq g \leq B. \quad (5)$$

Node propensity $\theta_i$ can be interpreted as a normalized degree of node $i$. In fact, the expected degree $\langle \tilde{d}_i \rangle$ of node $i$ is given by

$$\langle \tilde{d}_i \rangle = \left\langle \sum_{j=1}^{N} \tilde{w}_{ij} \right\rangle = \sum_{j=1}^{N} \lambda_{g_i,g_j} \theta_i \theta_j = \theta_i \sum_{g=1}^{B} \lambda_{g_i,g} \left( \sum_{j=1,g_j=g}^{N} \theta_j \right). \tag{6}$$

Because $\sum_{j=1,g_j=g}^{N} \theta_j = 1$ (Eq. (5)), we obtain

$$\theta_i = \frac{\langle \tilde{d}_i \rangle}{D_{g_i}}, \tag{7}$$

where $D_{g_i} = \sum_{g=1}^{B} \lambda_{g_i,g}$ is the number of edges emanating from nodes in group $g_i$, or equivalently, the sum of the degrees of nodes in group $g_i$.

For the graph with the expected edge weights, the transition probability from $i$ to $j$ is given by

$$P_{ij} = \frac{\langle \tilde{w}_{ij} \rangle}{\langle \tilde{d}_i \rangle} = \frac{\lambda_{g_i,g_j} \theta_i \theta_j}{\theta_i D_{g_i}} = \frac{\lambda_{g_i,g_j}}{D_{g_i}} \theta_j = P^{\text{SBM}}_{g_i,g_j} \theta_j, \tag{8}$$

where we define

$$P^{\text{SBM}}_{g,g'} := \lambda_{g,g'}/D_g, \tag{9}$$

by the fraction of edges to group $g'$ in $D_g$. Equation (8) makes clear that the transition of a walker from $i$ to $j$ in the dcSBM is decomposed into two random processes. The first is that a walker in group $g$ moves to group $g'$ with probability $P^{\text{SBM}}_{g,g'}$. The second is that, in group $g'$, the walker lands on node $j$ with probability $\theta_j$.

With this interpretation in mind, let us consider multi-step random walks. As a concrete example, let us consider the transition of a walker with two steps. The probability of moving from $i$ to $j$ by $t = 2$ steps via node $x$ is given by $P_{i,x} P_{x,j} = P^{\text{SBM}}_{g_i,g_x} P^{\text{SBM}}_{g_x,g_j} \theta_x \theta_j$. Summing over all nodes $x$ yields the probability of transiting from $i$ to $j$ with two steps:

$$\sum_{x=1}^{N} P_{i,x} P_{x,j} = \theta_j \sum_{x=1}^{N} P^{\text{SBM}}_{g_i,g_x} P^{\text{SBM}}_{g_x,g_j} \theta_x = \theta_j \sum_{g=1}^{B} P^{\text{SBM}}_{g_i,g} P^{\text{SBM}}_{g,g_j} \left( \sum_{x=1,g_x=g}^{N} \theta_x \right)$$

$$= \theta_j \sum_{g=1}^{B} P^{\text{SBM}}_{g_i,g} P^{\text{SBM}}_{g,g_j}$$

$$= \left( \mathbf{P}^2_{\text{SBM}} \right)_{g_i,g_j} \theta_j, \tag{10}$$

where $\mathbf{P}_{\text{SBM}} = (P^{\text{SBM}}_{ij})_{ij}$. Notice that the transition probability for the two-step random walk takes the same form with that for the one-step random walk (i.e., Eq. (8)), which is also true for the $t$-step random walk, i.e.,

$$\left( \mathbf{P}^t \right)_{ij} = \left( \mathbf{P}^t_{\text{SBM}} \right)_{g_i,g_j} \theta_j. \tag{11}$$

Therefore, the $t$-step random walks from node $i$ to $j$ can be described as two independent events, i.e., an event that the random walker moves from the block of node $i$ to the block of node $j$ by $t$ steps (i.e., $(\mathbf{P}^t_{\text{SBM}})_{g_i,g_j}$), and another event that the walker chooses node $j$ to visit among the nodes in the block $g_j$ (i.e., $\theta_j$).

3

The maximum likelihood estimation of dcSBM fits the parameters such that the degree of each node is preserved on average [8] by setting node propensity $\theta_i = d_i/D_{g_i}$ [8], which leads

$$\left(\mathbf{P}^t\right)_{i,j} = \frac{d_j}{D_{g_j}} \left(\mathbf{P}^t_{\text{SBM}}\right)_{g_i,g_j}. \tag{12}$$

Substituting Eq. (12) into Eq. (3) yields:

$$P_0(j \mid i) = \frac{d_j}{TD_j} \left(\sum_{t=1}^{T} \mathbf{P}^t_{\text{SBM}}\right)_{g_i,g_j}. \tag{13}$$

### 1.3 $P_0(j \mid i)$ for the special cases of the dcSBM

Let us derive $P_0(j \mid i)$ for two common null models, the Erdős-Rényi random graph [3] and the configuration model [4]. The Erdős-Rényi random graph is equivalent to the dcSBM with $B = 1$ and $\theta_i = 1/N$ [8], which gives

$$P_0(j \mid i) = 1/N. \tag{14}$$

The configuration model is equivalent to the dcSBM with $B = 1$ block and $\theta_i = d_i/2M$ [8], which gives

$$P_0(j \mid i) = d_j/2M. \tag{15}$$

### 1.4 $P_{\text{d}}(i) = P_0(i)$

The probability that the walker moves from node $i$ to a neighbor $j$ is given by

$$P_{\text{d}}(j \mid i) = \frac{1}{d_i}. \tag{16}$$

The random walk process in an undirected graph is ergodic, which ensures a unique stationarity for $t \to \infty$, with the detailed balance condition [10]:

$$P_{\text{d}}(j \mid i)P(i) = P_{\text{d}}(i \mid j)P_{\text{d}}(j) = c, \tag{17}$$

where $c > 0$ is a positive constant. By substituting Eq. (16) into Eq. (17), we have

$$P_{\text{d}}(i) = \frac{c}{P_{\text{d}}(j \mid i)} = cd_i. \tag{18}$$

Because $\sum_i P_{\text{d}}(i) = 1$, we obtain

$$P_{\text{d}}(i) = \frac{d_i}{\sum_\ell d_\ell}. \tag{19}$$

Notice that the stationary distribution of the random walker is proportional to degree $d_i$ and irrespective of the structure of the graph. Now, because both the dcSBM and the original graph have the same degree sequence on expectation, probability $P_0(i)$ for the dcSBM is given by

$$P_0(i) = \frac{d_i}{\sum_\ell d_\ell}. \tag{20}$$

4

## 1.5 Embedding directed graphs

Directed graphs may break the assumptions for the stationarity of random walks as well as the detailed balanced condition. Both assumptions are needed to compute $P_{\text{rw}}(x_{t-\ell} = j \mid x_t = i)$, i.e., a probability that the walker visits $j$ *before* visiting $i$. To avoid calculating the probability, we adopt a sliding window covering only the context nodes $j$ that appear *after* the center node $i$. This simplifies Eq. (1) into

$$P_{\text{d}}(j \mid i) = \frac{1}{T} \sum_{\tau=1}^{T} P_{\text{rw}}(x_{t+\tau} = j \mid x_t = i), \tag{21}$$

which in turn leads to the same expression as Eq. (3) without the assumptions of the stationarity and detailed balanced condition. A downside is that if node $i$ does not have outgoing edges (dangling node), then $i$ does not have any context $j$. To ensure that all nodes have at least one context, we allow a random walker to move against the direction of edges with a small probability $\epsilon = 0.05$. If a random walker hits any dangling node, it moves against a randomly selected in-coming edge.

## 2 Implementation of `residual2vec`

---
**Algorithm 1** Pseudocode of `residual2vec`.
---
**Input:**
    $\mathbf{A}$: Adjacency matrix of a graph of $N$ nodes
    $K$: Embedding dimension
    $\boldsymbol{g}$: Block membership of nodes for a null model (optional)
**Output:**
    Node embedding $\mathbf{U}$, where each $i$th row indicates the embedding of node $i$.
1: $\hat{\boldsymbol{g}} \leftarrow \text{FittingDCSBM}(\mathbf{A})$ //*Ref. [7]*
2: $\hat{\mathbf{R}} \leftarrow \text{TruncateResidualMatrix}(\mathbf{A}, \boldsymbol{g}, \hat{\boldsymbol{g}})$
3: $\boldsymbol{\lambda}, \mathbf{U} \leftarrow \text{RandomizedSVD}(\hat{\mathbf{R}}, K)$ //*Ref. [6]*
4: $\mathbf{U} \leftarrow \mathbf{U} \cdot \text{diag}(\sqrt{\boldsymbol{\lambda}})$
---

`residual2vec` has three components, namely i.e., (i) the block approximation, (ii) truncation, and (iii) matrix factorization (see Algorithm 1). We walk through how to implement each component efficiently in the following.

### 2.1 Block approximation

**Approximating $P_{\text{d}}(j \mid i)$ with the dcSBM** Our implementation centered on *the block approximation*, which substantially reduces the computational burden. Remind that `residual2vec` factorizes the residual matrix (Eq. (13) in the main text):

$$R_{ij} = \ln P_{\text{d}}(j \mid i) - \ln P_0(j \mid i). \tag{22}$$

Computing the second term (i.e., $\ln P_0(j \mid i)$) is easy because it can be computed by taking the power of a $B \times B$ matrix (i.e., Eq. (13)), and $B \ll N$ in practice.

Computing the other term $\ln P_{\text{d}}(j \mid i)$, however, is prohibitively difficult because it involves the power of $N \times N$ matrix, $\mathbf{P}$ (i.e., Eq. (3)), requiring the time and space

5

**Algorithm 2** TRUNCATERESIDUALMATRIX

**Input:**
  **A**: Adjacency matrix
  **g**: Group membership for random graphs
  **ĝ**: Group membership for the approximated graph
**Output:**
  **R̃**: Truncated residual matrix
1: $\hat{R}_{ij} \leftarrow 0$ //Initialize

  //Calculate the tentative $\tilde{R}_{ij}$ by truncating **L**
2: **for all** $i \in [1, N]$, $g \in [1, B]$ and $\hat{g} \in [1, \hat{B}]$ **do**
3:   **if** $h(i, g, \hat{g}) > 0$ **then**
4:     $\hat{R}_{ij} \leftarrow h(i, g, \hat{g})$ for all $g_j = g$ and $\hat{g}_j = \hat{g}$
5:   **end if**
6: **end for**

  //Re-evaluate Eq. (27) for $S_{ij} > 0$
7: **for all** the non-zero elements of **A do**
8:   $\hat{R}_{ij} \leftarrow \max(0, S_{ij} + h(i, g, \hat{g}))$
9: **end for**

---

complexities of $\mathcal{O}(N^3)$ and $\mathcal{O}(N^2)$, respectively. The block approximation remedies this problem by approximating the given graph with the dcSBM [8]. Specifically, we fit the dcSBM to the given graph using a maximum likelihood estimation [7]. We set the number $\hat{B}$ of groups to $\hat{B} = \min(N, 1000)$, where $N$ is the number of nodes. With Eq. (12), the transition matrix for the approximated graph, denoted by $\hat{\mathbf{P}} = (\hat{P}_{ij})$, is given by

$$\hat{P}_{ij} = \frac{d_j}{D_{g_j}} \hat{P}^{\text{SBM}}_{\hat{g}_i, \hat{g}_j}, \qquad (23)$$

where $\hat{g}_i$ is the group to which the block approximation assigns node $i$. The $t$th power $\hat{\mathbf{P}}^t$ can be computed by

$$\left(\hat{\mathbf{P}}^t\right)_{ij} = \frac{d_j}{D_{g_j}} \left(\hat{\mathbf{P}}^t_{\text{SBM}}\right)_{\hat{g}_i, \hat{g}_j}. \qquad (24)$$

It is the strength of this approximation that allows us to calculate the matrix power efficiently. Notice that $\hat{\mathbf{P}}^t$—which is the power of an $N \times N$ matrix—can be computed by $\hat{\mathbf{P}}^t_{\text{SBM}}$—which is the power of a smaller $\hat{B} \times \hat{B}$ matrix ($\hat{B} \leq N$). We take advantage of this property by using $\hat{\mathbf{P}}^t$ as the substitute of $\mathbf{P}^t$ for $t > 0$ in Eq. (3) to compute $\ln P_{\text{d}}(j \mid i)$, i.e.,

$$\ln P_{\text{d}}(j \mid i) = \ln \left[\frac{1}{T}\left(\sum_{t=1}^{T} \mathbf{P}^t\right)\right] \simeq \ln \left[\frac{1}{T}\left(\mathbf{P} + \mathbf{P} \sum_{t=1}^{T-1} \hat{\mathbf{P}}^t\right)\right]_{ij} \qquad (25)$$
$$=: \ln \hat{P}_{\text{d}}(j \mid i).$$

By substituting Eq. (24) into Eq. (25), we have

$$\ln \hat{P}_{\text{d}}(j \mid i) = \ln \frac{1}{T}\left[P_{ij} + \sum_{\ell=1}^{N} P_{i\ell} \left(\sum_{t=1}^{T-1} \hat{\mathbf{P}}^t_{\text{SBM}}\right)_{\hat{g}_\ell, \hat{g}_j} \frac{d_j}{D_{\hat{g}_j}}\right]. \qquad (26)$$

6

**Accuracy of the block approximation** The accuracy of the block approximation hinges on window size $T$ and the community structure of the graph. The block approximation is exact when $T = 1$ and $T = \infty$. For $1 < T < \infty$, the accuracy depends on how well the dcSBM approximates the given graph. The dcSBM describes the graph structure in terms of the connectivities of groups (or communities) and the degree of nodes. Therefore, the block approximation is a good approximation if the given graph has a strong community structure that is well described by the dcSBM.

As a proof of concept, we tested the block approximation using the empirical graphs listed in Table 2 in the main text (Fig. 1). We measured the Pearson correlation coefficient, denoted by $\rho$, between the exact and approximated $\ln P_{\mathrm{d}}(j \mid i)$. The airport network—which has a strong community structure that well fits the dcSBM [11]—has the largest correlation, $\rho$. By contrast, the correlation is relatively small for the DBLP citation graph. Yet, for all the graphs, the correlation is relatively high on average ($\rho = 0.81$). Overall, the correlation tends to increases as window size $T$ increases, suggesting that $P_{\mathrm{d}}(j \mid i)$ is well approximated, particularly for relatively large window size.

## 2.2 Truncation

`residual2vec` truncates the residual matrix by

$$\tilde{R}_{ij} = \max(0, R_{ij}), \qquad (27)$$

which costs time complexity of $\mathcal{O}(N^2)$ because $\mathbf{R} = (R_{ij})$ has $N^2$ elements. Here, we perform the truncation more efficiently by taking advantage of the block approximation.

The block approximation approximates the residual matrix by

$$\begin{aligned}
R_{ij} &= \ln \frac{1}{T} \left[ P_{ij} + \sum_{\ell=1}^{N} P_{i\ell} \left( \sum_{t=1}^{T-1} \hat{\mathbf{P}}_{\mathrm{SBM}}^{t} \right)_{\hat{g}_\ell, \hat{g}_j} \frac{d_j}{D_{\hat{g}_j}} \right] - \ln \frac{d_j}{TD_{g_j}} \left( \sum_{t=1}^{T} \mathbf{P}_{\mathrm{SBM}}^{t} \right)_{g_i, g_j} \\
&= \ln \left[ P_{ij} + \sum_{\ell=1}^{N} P_{i\ell} \left( \sum_{t=1}^{T-1} \hat{\mathbf{P}}_{\mathrm{SBM}}^{t} \right)_{\hat{g}_\ell, \hat{g}_j} \frac{d_j}{D_{\hat{g}_j}} \right] - \ln \frac{d_j}{D_{g_j}} \left( \sum_{t=1}^{T} \mathbf{P}_{\mathrm{SBM}}^{t} \right)_{g_i, g_j}.
\end{aligned} \qquad (28)$$

We note that $g_i$ and $\hat{g}_i$ are different, i.e., $g_i$ indicates the group of node $i$ for random graphs, and $\hat{g}_i$ indicates the group for approximating the given graph by the block approximation. The key consequence of the block approximation is that $\mathbf{R}$ can be decomposed into a block matrix, $\mathbf{L}$, and a sparse matrix $\mathbf{S}$, i.e.,

$$\mathbf{R} = \mathbf{L} + \mathbf{S}. \qquad (29)$$

By leveraging the nature of $\mathbf{L}$ and $\mathbf{S}$, the truncation can be performed efficiently in the following two steps. First, we calculate tentative $R_{ij}$ values by truncating $\mathbf{L}$. Each element of $\mathbf{L}$ takes one of a handful of $r$ ($r \ll N^2$) unique values. Therefore, $\mathbf{L}$—which consists of $N^2$ elements—can be truncated by truncating the $r$ values. Second, we go through each element $(i, j)$ for $S_{ij} > 0$ and re-evaluate Eq. (27), i.e., $\max(0, L_{ij} + S_{ij})$. Matrix $\mathbf{S}$ has $M$ non-zero elements, where $M$ is the number of edges. Therefore, $\mathbf{R}$ can be truncated by truncating $r + M$ values, which is a substantial reduction of computations compared to truncating all $N^2$ elements.

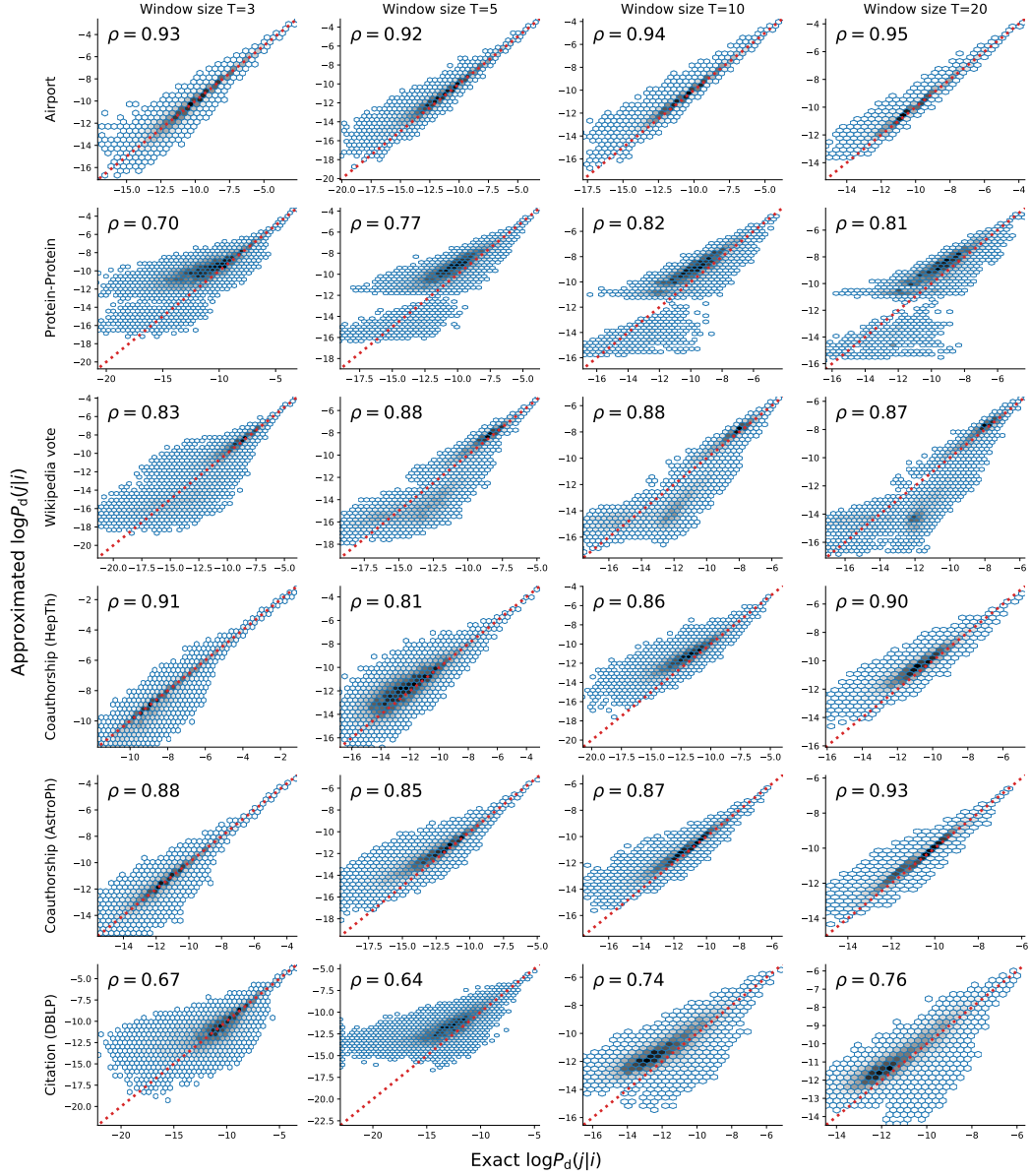

Figure 1: Approximated $P_{\mathrm{d}}(j \mid i)$ computed by the block approximation. The dashed diagonal line indicates the perfect match between the exact and approximated values. Variable $\rho$ in each panel indicates the Pearson correlation coefficient between the exact and approximated $\ln P_{\mathrm{d}}(j \mid i)$.

Specifically, **R** can be decomposed by

$$L_{ij} = \ln \sum_{\ell=1}^{N} P_{i\ell} \left( \sum_{t=1}^{T-1} \hat{\mathbf{P}}_{\text{SBM}}^{t} \right)_{\hat{g}_\ell, \hat{g}_j} - \ln \left( \sum_{t=1}^{T} \mathbf{P}_{\text{SBM}}^{t} \right)_{g_i, g_j} + \ln \frac{D_{g_j}}{D_{\hat{g}_j}}, \tag{30}$$

$$S_{ij} = \begin{cases} \ln \left( P_{ij} + \sum_{\ell=1}^{N} P_{i\ell} \left( \sum_{t=1}^{T-1} \hat{\mathbf{P}}_{\text{SBM}}^{t} \right)_{\hat{g}_\ell, \hat{g}_j} \right) \\ \quad - \ln \sum_{\ell=1}^{N} P_{i\ell} \left( \sum_{t=1}^{T-1} \hat{\mathbf{P}}_{\text{SBM}}^{t} \right)_{\hat{g}_\ell, \hat{g}_j} & (\text{if } P_{ij} > 0), \\ 0 & (\text{otherwise}). \end{cases} \tag{31}$$

Matrix **L** consists of at most $r = NB\hat{B}$ unique element values, i.e.,

$$L_{ij} \in \left\{ h(i, g, \hat{g}) \middle| i = 1, \ldots, N, g \in [1, B], \hat{g} \in [1, \hat{B}] \right\}, \tag{32}$$

where

$$h(i, g, \hat{g}) := \ln \sum_{\ell=1}^{N} P_{i\ell} \left( \sum_{t=1}^{T-1} \hat{\mathbf{P}}_{\text{SBM}}^{t} \right)_{\hat{g}_\ell, \hat{g}} - \ln \left( \sum_{t=1}^{T} \mathbf{P}_{\text{SBM}}^{t} \right)_{g_i, g} + \ln \frac{D_g}{D_{\hat{g}}}. \tag{33}$$

The pseudo code for the truncation is described in Table 2.

## 2.3 Matrix factorization

We factorize the truncated residual matrix, $\hat{\mathbf{R}}$, using the singular value decomposition (SVD). However, the SVD is practically infeasible for large graphs because its time complexity increases cubically with respect to $N$, i.e., $\mathcal{O}(N^3)$.

We circumvent this problem by leveraging the sparsity of $\hat{\mathbf{R}}$. The truncated residual matrix $\hat{\mathbf{R}}$ is sparse in our numerical simulations, e.g., at least 99% of the elements in **R** are zero for the six graphs. We take advantage of the sparsity of $\hat{\mathbf{R}}$ by using the randomized SVD (rSVD) [6]. The time and space complexities of the rSVD for computing $K$ leading eigenvectors are $\mathcal{O}((N + m)K)$ and $\mathcal{O}(NK)$, respectively, for an $N \times N$ sparse matrix with $m$ non-zero elements.

## 2.4 Computational complexity

The time and space complexities of each component in `residual2vec` are described in Table 1. We note that we used the rSVD for fitting the dcSBM instead of the SVD used in the original paper [7]. Taken together, the time complexity of `residual2vec` is $\mathcal{O}((N + M)\hat{B} + N\hat{B}^2 + T B^3 + M\hat{B} + T\hat{B}^3 + M + NB\hat{B} + \text{nnz}(N + \hat{\mathbf{R}})K + NK) = \mathcal{O}((N + M)\hat{B} + T\hat{B}^3)$, where we have assumed $B, K \leq \hat{B} \leq N$ and $\mathcal{O}(M) = \mathcal{O}(\text{nnz}(\hat{\mathbf{R}}))$. The space complexity is $\mathcal{O}(N\hat{B} + B^2 + N\hat{B} + B + \hat{B} + NB\hat{B} + NK) = \mathcal{O}(NB\hat{B})$.

# 3 Supplementary information for the experiments

## 3.1 Offset $z_i$ of `residual2vec` for link prediction

`residual2vec` decomposes node similarities into two components, i.e., embedding $u_i$ and baseline probability $P_0(j|i)$ in Eq. (10) in the main text. Baseline probability $P_0(j|i)$

9

Table 1: Time and space complexities of `residual2vec`.

| | | Complexity | |
|---|---|---|---|
| | Process description | Time | Space |
| Block approximation | Compute the $\hat{B}$ leading vectors of the adjacency matrix of the graph with $N$ nodes and $M$ edges using the rSVD [6]. | $\mathcal{O}((N+M)\hat{B} + \hat{B}^3)$ | $\mathcal{O}(N\hat{B})$ |
| | Perform the $K$-means clustering for $N$ nodes with $\hat{B}$ dimensional vectors [1]. | $\mathcal{O}(N\hat{B}^2)$ | $\mathcal{O}(N\hat{B})$ |
| Truncation | $\sum_{t=1}^{T} \mathbf{P}_{\text{SBM}}^t$ | $\mathcal{O}(TB^3)$ | $\mathcal{O}(B^2)$ |
| | $\sum_{\ell=1}^{N} P_{i\ell} \left( \sum_{t=1}^{T} \hat{\mathbf{P}}_{\text{SBM}}^t \right)_{g_\ell, g}, \forall i \in [1, N]$ and $g \in [1, \hat{B}]$. | $\mathcal{O}(M\hat{B} + T\hat{B}^3)$ | $\mathcal{O}(N\hat{B})$ |
| | $D_g, \forall g \in [1, B]$ | $\mathcal{O}(M)$ | $\mathcal{O}(B)$ |
| | $D_{\hat{g}}, \forall g \in [1, \hat{B}]$ | $\mathcal{O}(M)$ | $\mathcal{O}(\hat{B})$ |
| | Eq. (32) | $\mathcal{O}(NB\hat{B})$ | $\mathcal{O}(NB\hat{B})$ |
| Matrix factorization | Compute the $K$ leading vectors of $\hat{\mathbf{R}}$ using the rSVD [6]. | $\mathcal{O}(\text{nnz}((\hat{\mathbf{R}}) + N)K)$ | $\mathcal{O}(NK)$ |
| | Scaling the dimensions by singular values | $\mathcal{O}(NK)$ | $\mathcal{O}(NK)$ |

accounts for the similarities attributed to a null model, and embedding similarity $u_i^\top u_j$ represents the "residual" from baseline probability $P_0(j|i)$. In the link prediction task, we aimed to leverage both baseline and residual similarities for prediction, by adding offset $z_j = \ln P_0(j|i)$ to the embedding similarity $u_i^\top u_j$. We added $\ln P_0$ instead of $P_0$ by noting that Eq. (10) in the main text can be rewritten as

$$P_{\text{r2v}}(j|i) = \frac{P_0(j|i) \exp(u_i^\top u_j)}{Z_i'} = \frac{\exp(u_i^\top u_j + \ln P_0(j|i))}{Z_i'}. \tag{34}$$

In other words, $\ln P_0(j|i)$ has the same unit as the embedding similarity $u_i^\top u_j$ in the model. Therefore, we adopted $\ln P_0(j|i)$ as the offset $z_i$.

## 3.2 Linear regression model for node similarities

In the Case study section, we quantified the strength of bias due to degree and time using a linear regression model $y_{ij} = f(x_i, x_j)$ that explains cosine similarity $y_{ij}$ for node pair $(i, j)$ by the degrees and the years of nodes $i$ and $j$, i.e., $x_i$ and $x_j$. A simple linear regression model is given by

$$\hat{y}_{ij} = w_0 + w_1 x_i + w_2 x_j, \tag{35}$$

Because the embedding similarity is symmetric ($y_{ij} = y_{ji}$), the modeled similarities should be also symmetric, i.e., $\hat{y}_{ij} = \hat{y}_{ji}$. By substituting Eq. (35) into $\hat{y}_{ij} = \hat{y}_{ji}$ yields $w_1 = w_2$, which leads

$$y_{ij} = w_0 + w_1(x_i + x_j). \tag{36}$$

This model explains the similarity by the sum of the node features, $x_i + x_j$. Instead of the summation, one can use the difference $|x_i - x_j|$ or product $x_i x_j$ for predicting the similarity $y_{ij}$. Here, we put them together into one linear regression model $\hat{y}_{ij} = w_0 + w_1(x_i + x_j) + w_2|x_i - x_j| + w_3 x_i x_j$. This composite model contains more variables and thus would explain the similarity $y_{ij}$ better than any of the model that only uses one variable.

10

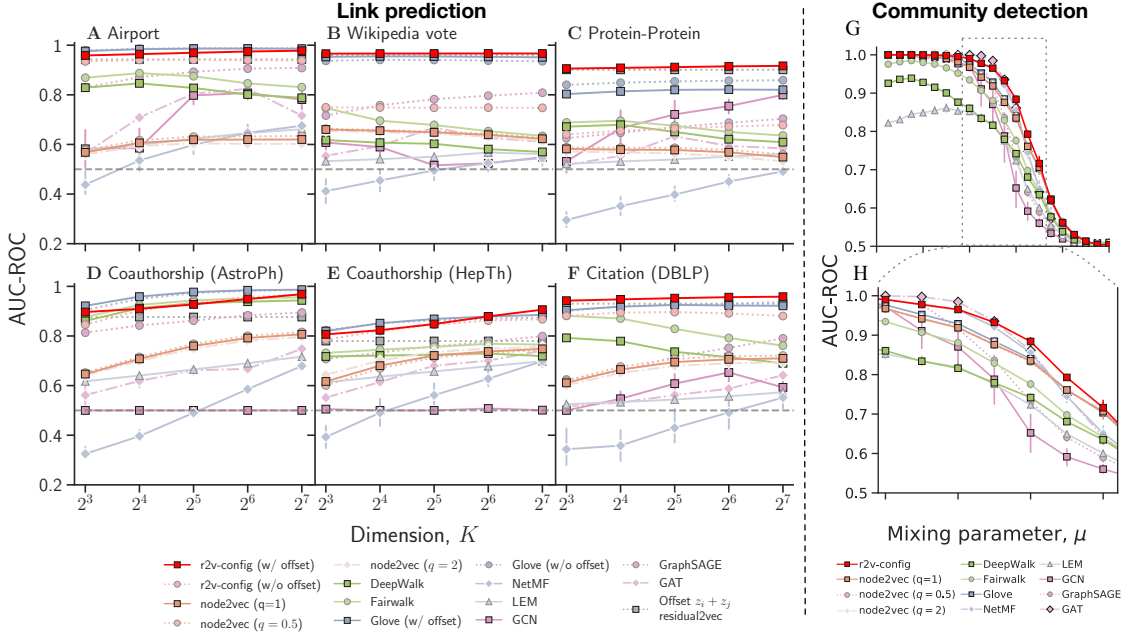

Figure 2: (A)—(F) The full results for the link prediction benchmark. (G–H) The results for the LFR model with homogeneous degree distributions. The power-law exponent for the degree distribution is set to $\gamma = 6$.

### 3.3 Full results for the link prediction benchmark

The full results for the link prediction benchmark are shown in Figs. 2A—F.

### 3.4 Community detection benchmark for graphs with homogeneous degree distributions

We performed the community detection benchmark with graphs having homogeneous degree distribution. Specifically, we generated the graphs using the LFR benchmark with the exponent of the degree distribution set to $\gamma = 6$. Other configurations for the benchmark are the same as those described in the main text. The results for the benchmark are shown in Figs. 2G and H.

### 3.5 GCN is not sensitive to the dimension of node features

In the benchmark, we use the eigenvectors associated with the smallest eigenvalues of the normalized Laplacian matrix as the input node features for `GCN`, `GraphSAGE`, and `GAT`. Although we used $\hat{K} = K$ eigenvectors, one can input more eigenvectors. Figure 3 shows the benchmark performance of the GCN and GraphSAGE with $\hat{K} = K$ or $\hat{K} = 2K$ eigenvectors as the input. Even if we input more eigenvectors ($\hat{K} = 2K$), the performance does not increase much or even gets worse. This may be because, for the link prediction and community detection tasks, $\hat{K} = K$ eigenvectors are sufficient. Furthermore, the more features we input, the more parameters the models have, making it difficult for the optimization algorithm to find a good parameter set.

11

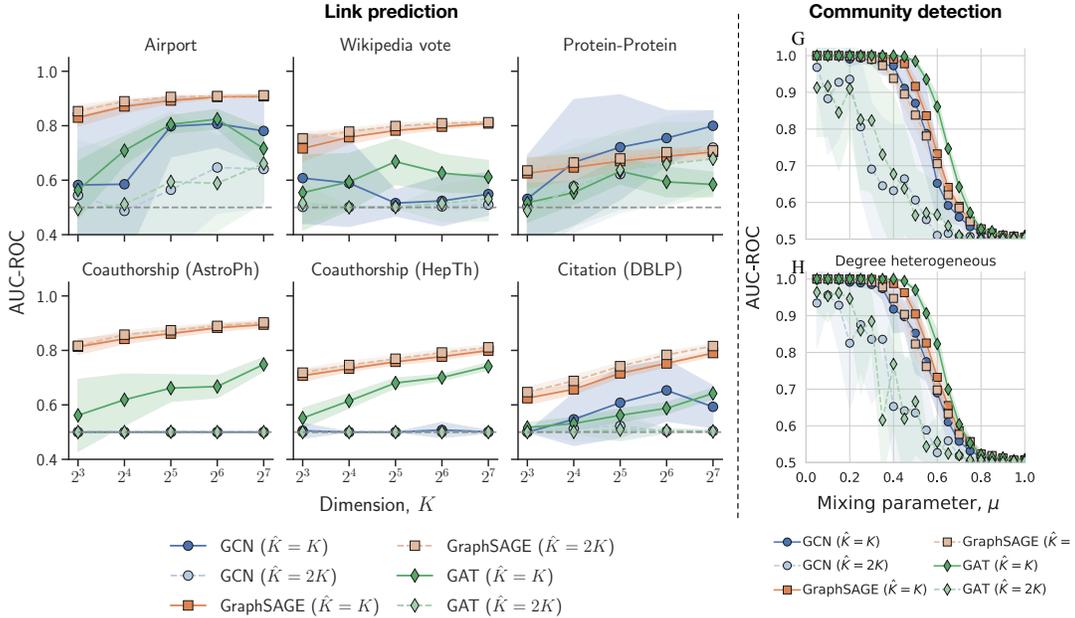

Figure 3: Performance for the link prediction and community detection benchmarks.

### 3.6 Results of directed citation graph of journals

Citations are inherently directional. However, we neglected the directionality and embedded the undirected citation graph of journals. This is because `residual2vec` and the baseline methods perform worse for the directed graphs in terms of the prediction of the impact and subject category of journals.

Figures 4 A and B show the 2D projection of the embedding generated by the Linear Discriminant Analysis, with the subject categories being the class labels. As is the case for the undirected graphs, `Glove` and `node2vec` strongly capture the temporal information. By contrast, `r2v-dcSBM` better delineates the subject categories more clearly than `Glove` and `node2vec`. In the embedding generated by `r2v-config`, node similarity—measured by the cosine similarity of the embedding vectors—is relatively independent of the degree and year compared to the embedding generated by `Glove`, `node2vec`, and `r2v-dcSBM` (Fig. 4C).

We predict the journals' impact factor and subject categories using the $k$-means algorithm with 5-cross validations. Although all methods predict worse compared to the embedding of the undirected graph, `r2v-config` best predicts the disciplines and impact of journals (Fig. 4D).

### 3.7 Web of Science citation graph

We construct a citation graph of journals using the citation data taken from the Web of Science (WoS) in 2020. The dataset contains bibliographic information including $1,547,459,602$ citations among $496,833,161$ papers published from $28,495$ journals in years between 1900 and 2019 in various research fields. We retrieve the subject category of each journal from the three journal collections, i.e., Science Citation Index Expanded (SCIE), Social Sciences Citation Index (SSCI), and Arts & Humanities Citation Index (AHCI). Each collection has a different category scheme. If a journal is indexed in mul-

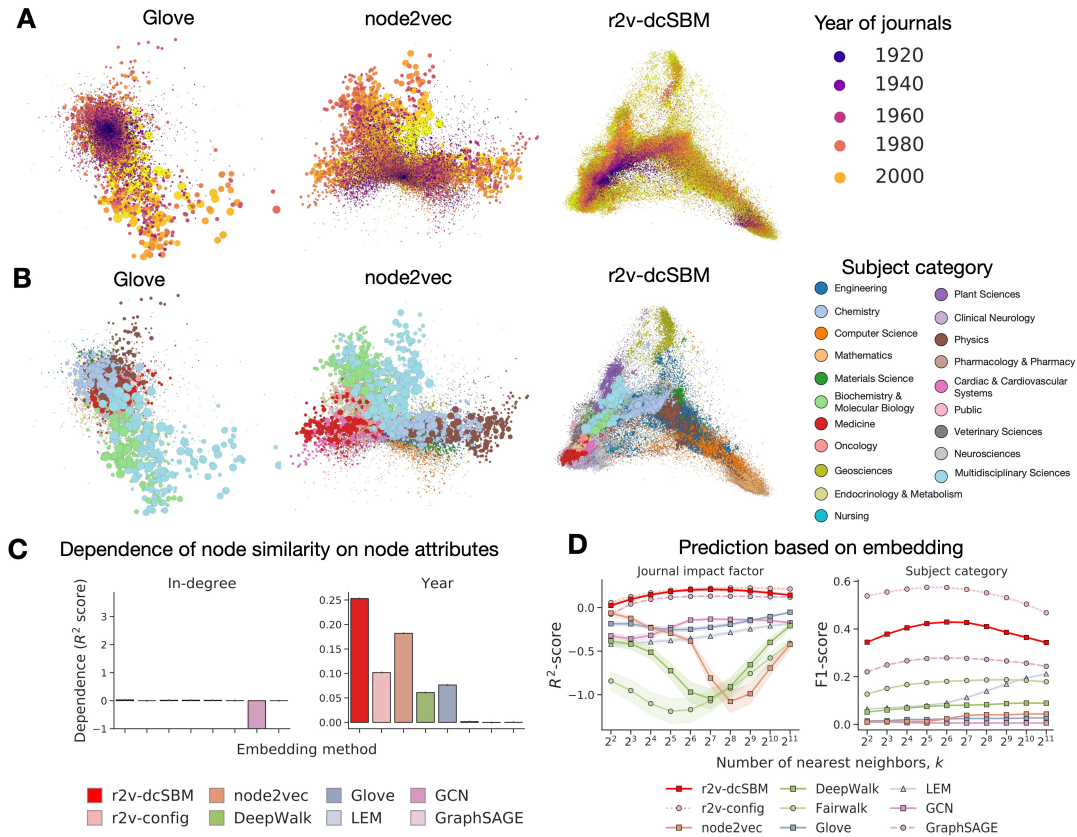

Figure 4: Embedding of the directed journal citation graph constructed from the WoS. (**A**, **B**) A 2d projection of 128-dimensional embedding by the Linear Discriminant Analysis. (**C**) Dependence of node similarity on nodes' degree and year. (**D**) By using the k-nearest neighbor algorithm, the embedding by `r2v-config` best predicts the impact factor and subject category.

tiple collections, we choose the largest collection and use its subject category for the journal. If the journal has multiple subjects within the chosen collection, we go through the table of the collection from the first to the last rows and use the one that first appears in the table.

## 3.8 Code

We implemented `DeepWalk` and `node2vec` using gensim package [12], with the same parameters used in the paper of node2vec [5]. We used `Glove` implemented in glove-python package [9]. We used `GCN` and `GraphSAGE` implemented in StellarGraph [2] and trained them using negative sampling.

## 3.9 Hardware

We ran experiments using a single machine with 64 Intel(R) Xeon(R) Gold 5218 CPU and 1Tb of RAM.



\end{document}